\documentstyle[aclap,epsfig]{article}

\title{\vspace{-0.5in}A Structured Language Model}

\author{Ciprian Chelba \\
The Johns Hopkins University\\
CLSP, Barton Hall 320\\
3400 N. Charles Street, Baltimore, MD-21218\\
\verb+chelba@jhu.edu+\\}
\begin{document}

\maketitle
\vspace{-0.5in}
\begin{abstract}

The paper presents a language model that develops syntactic structure
and uses it to extract meaningful information from the word
history, thus enabling the use of long distance dependencies. The
model assigns probability to every joint sequence of
words--binary-parse-structure with headword annotation. 
The model, its probabilistic parametrization, and a set of experiments
meant to evaluate its predictive power are presented.

\end{abstract}

\section{Introduction}

The main goal of the proposed project is to develop a language model(LM)
that uses syntactic structure.
The principles that guided this proposal were: \\
\hspace*{.3cm}$\bullet$ the model will develop syntactic knowledge as a built-in feature;
  it will assign a probability to every joint sequence of
  words--binary-parse-structure;\\
\hspace*{.3cm}$\bullet$ the model should operate in a left-to-right manner so that it would be possible
  to decode word lattices provided by an automatic  speech recognizer.\\
The model consists of two modules: a next word \emph{predictor} which makes
use of syntactic structure as developed by a \emph{parser}. The
operations of these two modules are intertwined. 

\section{The Basic Idea and Terminology}

\begin{figure}
  \begin{center} 
    \epsfig{file=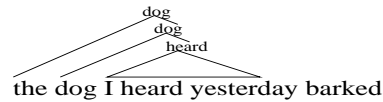,height=1.3cm,width=5cm}
  \end{center}
  \caption{Partial parse} \label{fig:p_parse}
\end{figure}

\begin{figure} 
  \begin{center}
    \epsfig{file=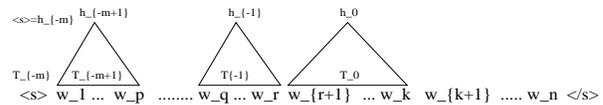,height=1.3cm,width=\columnwidth}
  \end{center}
  \caption{A word-parse k-prefix} \label{fig:w_parse}
\end{figure}

\begin{figure} 
  \begin{center} 
    \epsfig{file=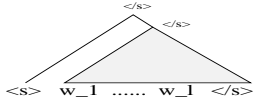,height=1.3cm,width=3.5cm}
  \end{center}
  \caption{Complete parse} \label{fig:c_parse}
\end{figure}

Consider predicting the word \verb+barked+ in the sentence:
\hspace*{.3cm}\verb+the dog I heard yesterday barked again.+\\
A 3-gram approach would predict \verb+barked+ from 
\verb+(heard, yesterday)+ whereas it is clear that the predictor should  use the
word \verb+dog+ which is outside the reach of even 4-grams. Our
assumption is that what enables us to make a good prediction of
\verb+barked+ is the syntactic structure in the past. 
The correct \emph{partial parse} of the word history when predicting
\verb+barked+ is shown in Figure~\ref{fig:p_parse}. 
The word \verb+dog+ is called the \emph{headword} of the
\emph{constituent} \verb+( the (dog (...)))+
and \verb+dog+ is an \emph{exposed headword} when predicting \verb+barked+ --- topmost headword in the
largest constituent that contains it. The syntactic structure in the past filters out
irrelevant words and points to the important ones, thus enabling the
use of long distance information when predicting the next word. 
Our model will assign a probability $P(W,T)$ to every sentence $W$
with every possible binary branching parse $T$ and every possible headword
annotation for every constituent of $T$.  
Let $W$ be a sentence of length $l$ words to which we have prepended
\verb+<s>+ and appended \verb+</s>+ so that $w_0 = $\verb+<s>+ and
$w_{l+1} = $\verb+</s>+.
Let $W_k$ be the word k-prefix $w_0 \ldots w_k$ of the sentence and 
\mbox{$W_k T_k$} the \emph{word-parse k-prefix}. To stress this point, a
\mbox{word-parse k-prefix} contains only those binary
trees whose span is completely included in the word k-prefix, excluding 
$w_0 = $\verb+<s>+. Single words can be regarded as root-only
trees. Figure~\ref{fig:w_parse} shows a word-parse k-prefix; \verb|h_0
.. h_{-m}| are the \emph{exposed headwords}. 
A \emph{complete parse} --- Figure~\ref{fig:c_parse} --- is any
binary parse of the \mbox{$w_1 \ldots w_l$\verb+ </s>+}
sequence with the restriction that \verb+</s>+ is the only  allowed
headword. Note that \mbox{$(w_1 \ldots w_l)$} \emph{needn't} be a
constituent, but for the parses where it is, there is no restriction on
which of its words is the headword.

The model will operate by means of two modules:\\
\hspace*{.3cm}$\bullet$ PREDICTOR predicts the next word $w_{k+1}$ given the word-parse
  k-prefix and then passes control to the PARSER;\\
\hspace*{.3cm}$\bullet$ PARSER grows the already existing binary branching structure by
  repeatedly generating the transitions \verb+adjoin-left+ or
  \verb+adjoin-right+ until it passes control to the PREDICTOR by
  taking a \verb+null+ transition.

\begin{figure}
  \begin{center} 
    \epsfig{file=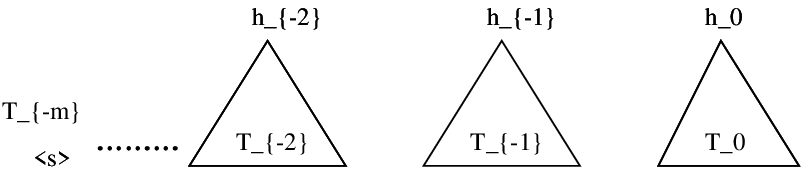,height=1.3cm,width=\columnwidth}
  \end{center}
  \caption{Before an adjoin operation} \label{fig:before}
  \begin{center} 
    \epsfig{file=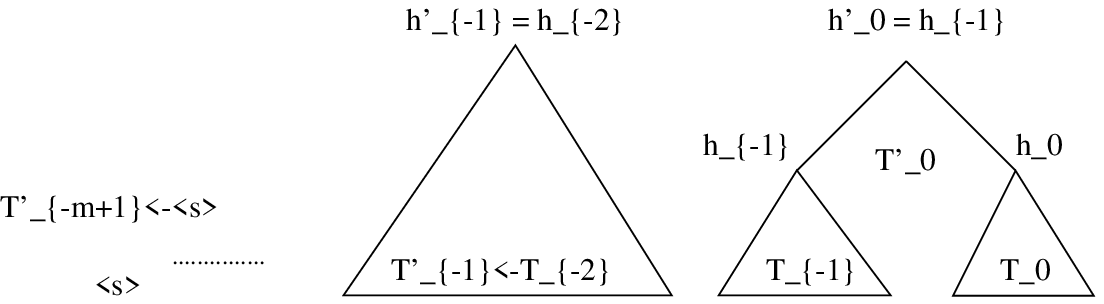,height=1.3cm,width=\columnwidth}
  \end{center}
  \caption{Result of adjoin-left} \label{fig:after_a_l}
  \begin{center} 
    \epsfig{file=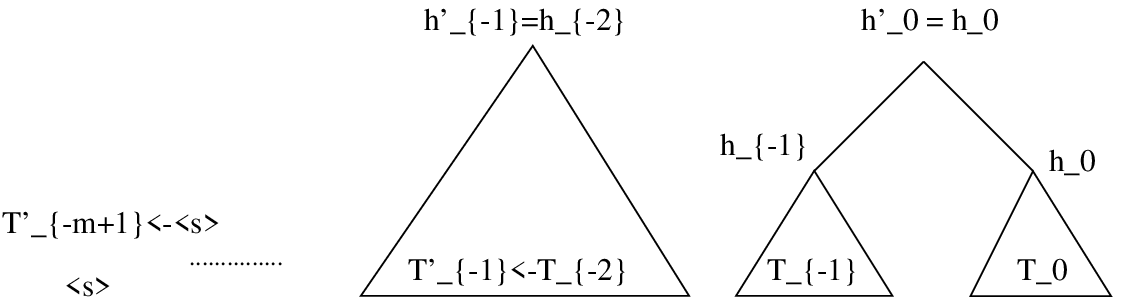,height=1.3cm,width=\columnwidth}
  \end{center}
  \caption{Result of adjoin-right} \label{fig:after_a_r}
\end{figure}

The operations performed by the PARSER ensure that all
possible binary branching parses with all possible headword
assignments for the $w_1 \ldots w_k$ word sequence can be generated.
They are illustrated by Figures~\ref{fig:before}-\ref{fig:after_a_r}.
The following algorithm describes how the model generates a word
sequence with a complete parse (see Figures~\ref{fig:c_parse}-\ref{fig:after_a_r} for notation):
\begin{verbatim}
Transition t;         // a PARSER transition
generate <s>;
do{
  predict next_word;             //PREDICTOR
  do{                            //PARSER
    if(T_{-1} != <s> )
       if(h_0 == </s>)     t = adjoin-right;
       else t = {adjoin-{left,right}, null};
    else    t = null;
  }while(t != null)
}while(!(h_0 == </s> && T_{-1} == <s>))
t = adjoin-right; // adjoin <s>; DONE
\end{verbatim}
It is easy to see that any given word sequence with a possible parse
and headword annotation is generated by a unique sequence of
model actions.

\section{Probabilistic Model}

The probability $P(W,T)$ can be broken into:\\
\mbox{ $P(W,T)=\prod_{k=1}^{l+1}[P(w_k/W_{k-1}T_{k-1}) \cdot$ }
\mbox{ $\prod_{i=1}^{N_k}P(t_i^k/w_k,W_{k-1}T_{k-1},t_1^k\ldots t_{i-1}^k)]$ }
where: \\
\hspace*{.3cm}$\bullet$ $W_{k-1} T_{k-1}$ is the word-parse $(k-1)$-prefix\\
\hspace*{.3cm}$\bullet$ $w_k$ is the word predicted by PREDICTOR\\
\hspace*{.3cm}$\bullet$ $N_k -1$ is the number of adjoin operations the PARSER executes
  before passing control to the  PREDICTOR (the $N_k$-th operation at
  position k is the \verb+null+ transition); $N_k$ is a function of $T$\\
\hspace*{.3cm}$\bullet$ $t_i^k$ denotes the i-th PARSER operation carried out at
  position k in the word string; \\ \mbox{ $t_{i}^k \in \{$\verb+adjoin-left+$,
  $\verb+adjoin-right+$\}, i < N_k$ }, \mbox{ $t_{i}^k =
  $\verb+null+$, i = N_k$ } \\
Our model is based on two probabilities:
\begin{eqnarray}
  P(w_k/W_{k-1} T_{k-1})\label{eq:1}\\
  P(t_i^k/w_k,W_{k-1} T_{k-1}, t_1^k \ldots t_{i-1}^k)\label{eq:2}
\end{eqnarray}

As can be seen \mbox{$(w_k,W_{k-1} T_{k-1}, t_1^k \ldots
  t_{i-1}^k)$} is one of the $N_k$ word-parse k-prefixes of $W_k T_k,
i=\overline{1,N_k}$ at position $k$ in the sentence. 

To ensure a proper probabilistic model we have to make sure
that~(\ref{eq:1}) and~(\ref{eq:2}) are well defined conditional
probabilities and that the model halts with probability one. A few
provisions need to be taken: \\
\hspace*{.3cm}$\bullet$  $P($\verb+null+$/W_k T_k) = 1$, if \verb+T_{-1} == <s>+
    ensures that \verb+<s>+ is adjoined in the last step of the
    parsing process;\\
\hspace*{.3cm}$\bullet$  $P($\verb+adjoin-right+$/W_k T_k) = 1$, if \verb+h_0 == </s>+
    ensures that the headword of a complete parse is \verb+</s>+;\\
\hspace*{.3cm}\mbox{$\bullet \exists\epsilon > 0 s.t.\
  P(w_k$=\verb+</s>+$/W_{k-1}T_{k-1})\geq\epsilon, \forall W_{k-1}T_{k-1}$} ensures that the model halts with probability one.\\

\subsection{The first model}

The first term~(\ref{eq:1}) can be reduced to an $n$-gram LM,
\mbox{$P(w_k/W_{k-1} T_{k-1}) = P(w_k/w_{k-1} \ldots w_{k-n+1})$}.

A simple alternative to this degenerate approach would be to build a model which
predicts the next word based on the preceding p-1 \emph{exposed headwords}
and n-1 words in the history, thus making the following equivalence classification:\\
$[W_k T_k] = \{$\verb|h_0 .. h_{-p+2},|$w_{k-1} .. w_{k-n+1}\}$.\\ The
approach is similar to the trigger LM\cite{Roni}, the difference being
that in the present work triggers are identified using the syntactic
structure.

\subsection{The second model}

Model~(\ref{eq:2}) assigns probability to different binary parses of the word
k-prefix by chaining the elementary operations described above. 
The workings of the PARSER are very similar to those of Spatter~\cite{Fred}.
It can be brought to the full power of Spatter by changing the action
of the \verb+adjoin+ operation so that it takes into account the
terminal/nonterminal labels of the constituent proposed by
\verb+adjoin+ and it also predicts the nonterminal label of the newly created constituent; PREDICTOR will now predict the next word along with its POS tag.
The best equivalence classification of the \mbox{$W_k T_k$} word-parse
k-prefix is yet to be determined. The Collins parser~\cite{Mike} shows that
dependency-grammar--like bigram constraints may be the most adequate,
so the equivalence classification $[W_k T_k]$ should contain at least
\verb|{h_0, h_{-1}}|.

\section{Preliminary Experiments}

Assuming that the correct partial parse is a function of the
word prefix, it makes sense to compare the word level perplexity(PP) of a
standard n-gram LM with that of the $P(w_k/W_{k-1}T_{k-1})$ model. 
We developed and evaluated four LMs: \\
\hspace*{.3cm}$\bullet$ 2 bigram LMs $P(w_k/W_{k-1}T_{k-1})=P(w_k/w_{k-1})$ referred
to as W and w, respectively; $w_{k-1}$ is the previous (word, POStag) pair;\\
\hspace*{.3cm}$\bullet$ 2 $P(w_k/W_{k-1}T_{k-1})=P(w_k/h_{0})$ models, referred to
as H and h, respectively; ${h_0}$ is the previous exposed (headword, POS/non-term tag) pair;
the parses used in this model were those assigned manually in the Penn
Treebank~\cite{Upenn} after undergoing headword percolation and binarization.

All four LMs predict a word $w_k$ and they were implemented using the
Maximum Entropy Modeling Toolkit\footnote{ftp://ftp.cs.princeton.edu/pub/packages/memt}~\cite{Eric}.
The constraint templates in the \{W,H\} models were:\\
\verb+  4 <= <*>_<*> <?>; 2 <= <?>_<*> <?>;+\\
\verb+  2 <= <?>_<?> <?>; 8 <= <*>_<?> <?>;+\\
and in the \{w,h\} models they were:\\
\verb+  4 <= <*>_<*> <?>; 2 <= <?>_<*> <?>;+
\verb+<*>+ denotes a {\em don't care} position, \verb+<?>_<?>+ a
(word, tag) pair; for example, \verb+4 <= <?>_<*> <?>+ will trigger on
all ((word, {\em any tag}), predicted-word) pairs that occur more than
3 times in the training data.  The sentence boundary is not included
in the PP calculation. Table~\ref{table:results} shows the PP results
along with the number of parameters  for each of the 4 models described .
\begin{center}
  \begin{table}[h]
     \begin{center}
       \begin{tabular}{||c|c|c||c|c|c||}                           \hline
         LM      &   PP   & param   &  LM      &   PP   & param  \\ \hline \hline 
         W       &   352  & 208487  &  w       &   419  & 103732 \\ \hline
         H       &   292  & 206540  &  h       &   410  & 102437  \\ \hline
       \end{tabular}
     \end{center}
     \caption{Perplexity results} \label{table:results}
   \end{table}
\end{center}
\section{Acknowledgements}

The author thanks to all the members of the Dependency Modeling
Group~\cite{ws96}:David Engle, Frederick Jelinek, Victor Jimenez,
Sanjeev Khudanpur, Lidia Mangu, Harry Printz, Eric Ristad, Roni
Rosenfeld, Andreas Stolcke, Dekai Wu.

\end{document}